\documentclass[10pt,letterpaper]{article}

\usepackage{cogsci}

\cogscifinalcopy 

\usepackage[
    backend=biber,
    style=apa,
    natbib=true,
    doi=false,
    isbn=false,
    url=false,
]{biblatex}
\usepackage{pslatex}
\usepackage{float} 

\usepackage{graphicx}
\usepackage{multirow}
\usepackage{xspace}
\usepackage{hyperref}
\usepackage[nameinlink]{cleveref}

\newcommand{\aautoref}[1]{\hyperref[#1]{Appendix~\ref*{#1}}}

\usepackage{booktabs}
\usepackage[dvipsnames]{xcolor}

\newcommand{\eat}[1]{}

\newcommand{\kilogram}{\textsc{KiloGram}\xspace}

\title{Semantic uncertainty guides the extension of conventions to new referents}

 \author{{\large \bf Ron Eliav\textsuperscript{1},
 Anya Ji\textsuperscript{2},
 Yoav Artzi\textsuperscript{2}, 
 Robert D. Hawkins\textsuperscript{3*}} \\
\textsuperscript{1} Department of Computer Science, Bar-Ilan University \\
\textsuperscript{2}Department of Computer Science, Cornell University \\
\textsuperscript{3}Princeton Neuroscience Institute, Princeton University \\ \textsuperscript{*}Correspondence to \texttt{rdhawkins@princeton.edu}
}
\begin{document}

\maketitle

\begin{abstract}

A long tradition of studies in psycholinguistics has examined the formation and generalization of \emph{ad hoc} conventions in reference games, showing how newly acquired conventions for a given target transfer to new referential contexts. 
However, another axis of generalization remains understudied: how do conventions formed for one target transfer to \emph{completely distinct targets}, when specific lexical choices are unlikely to repeat? 
This paper presents two dyadic studies $(N=240$) that address this axis of generalization, focusing on the role of nameability --- the \emph{a priori} likelihood that two individuals will share the same label. 
We leverage the recently-released \kilogram dataset, a collection of abstract tangram images that is orders of magnitude larger than previously available, exhibiting high diversity of properties like nameability. 
Our first study asks how nameability shapes convention formation, while the second asks how new conventions generalize to entirely new targets of reference. 
Our results raise new questions about how \emph{ad hoc} conventions extend beyond target-specific re-use of specific lexical choices. 

\end{abstract}

\section{Introduction}

A core problem that all theories of reference must address is the problem of \emph{generalization}. 
When faced with a novel referent, we must somehow extend or combine existing items from our lexicon to produce (as a speaker) or interpret (as a listener) an utterance with the new meaning.
This problem was sidestepped under classical truth-conditional approaches to reference, where the exhaustive set of referents an utterance applied to was built into its literal meaning \citep{lewis1976general,dale1995computational}. 
More recently, however, a family of probabilistic approaches to reference have emerged to handle graded phenomena like vagueness~\citep{goodman2015probabilistic} and typicality~\citep{degen2020redundancy}.
Under these accounts, interlocutors maintain graded \emph{uncertainty} about exactly what an expression will mean to others~\citep{potts2015embedded,bergen2016pragmatic}.
Hence, referential generalization is inherently risky.
There is no guarantee that a given meaning will be shared by one's partner, and the reference may fail.

\begin{figure}[t!]
    \centering
    \fbox{\includegraphics[keepaspectratio=true, height=4.5em]{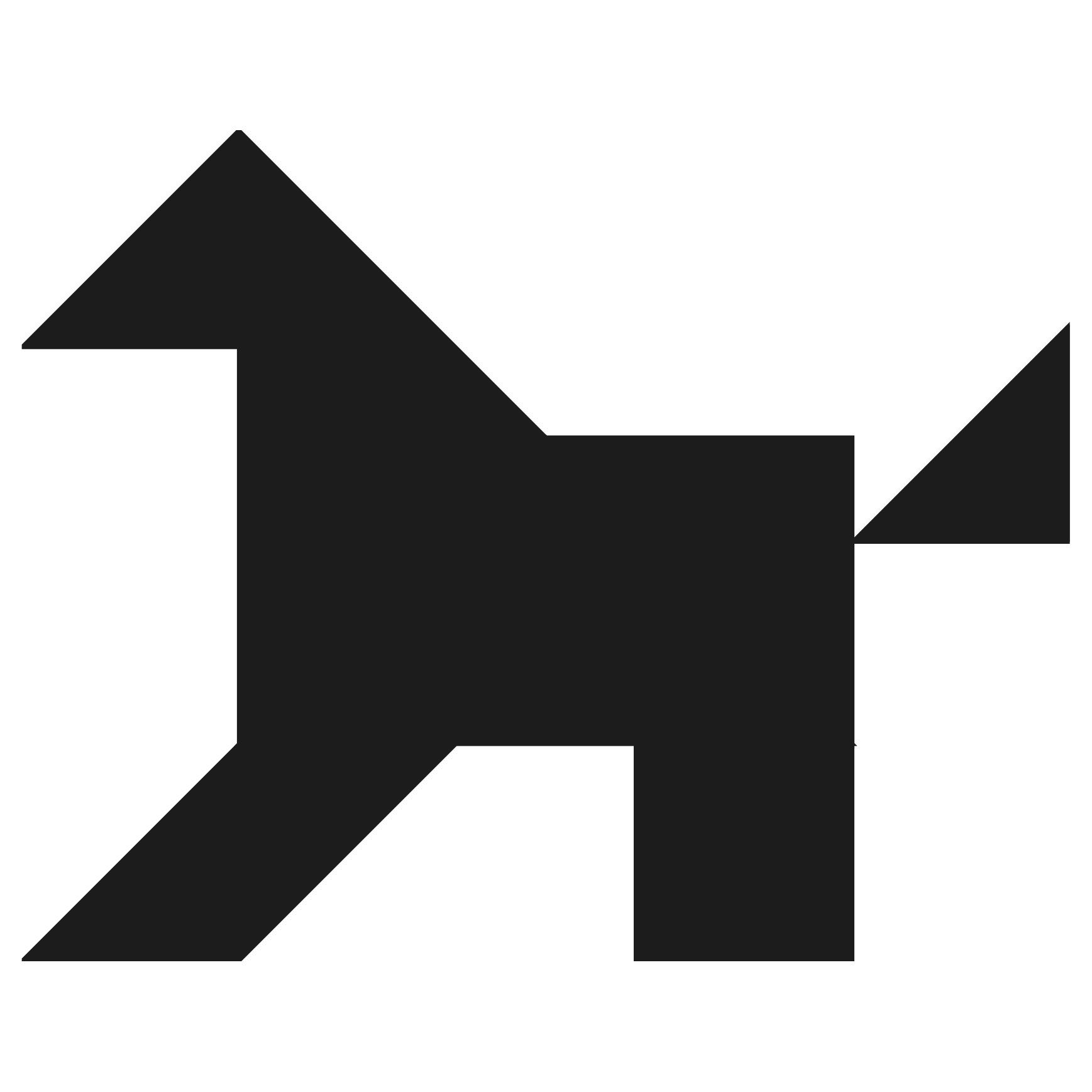}}\hspace{2em}\fbox{\includegraphics[keepaspectratio=true, height=4.5em]{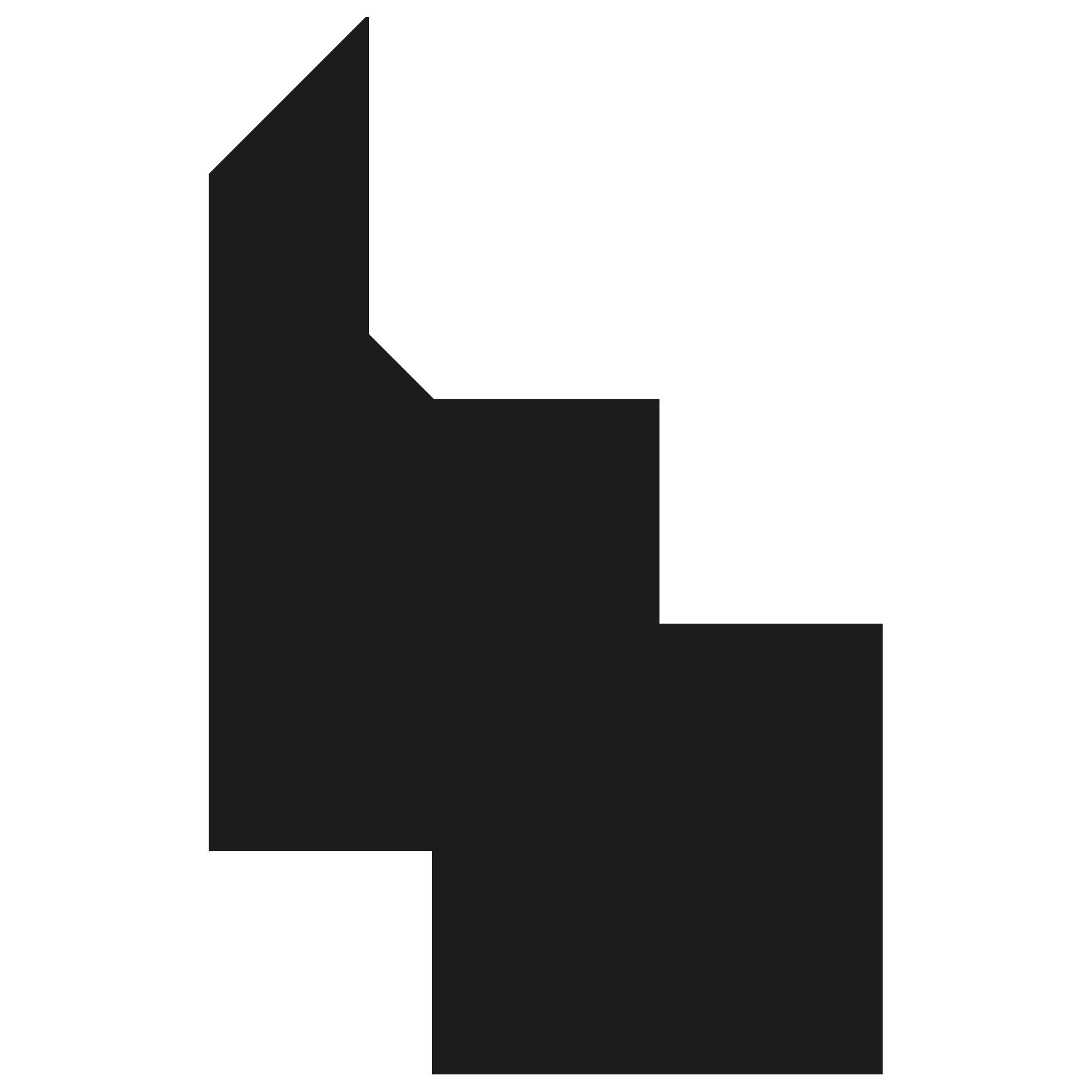}}
    \caption{Example tangram images from the \kilogram dataset. The one on the left has relatively high nameability, and the one on the right lower.}
    \label{fig:introtangrams}
\end{figure}

How, then, do interlocutors decide which expressions would be more or less effective when faced with something new?
One important source of evidence comes from a classic line of work using repeated reference games with deliberately ambiguous referential targets like abstract drawings or tangram shapes~\citep{krauss1964changes,clark1986referring,hawkins2020characterizing}.
These studies reliably establish the basic conditions for studying referential generalization: participants must rely on their pre-existing lexical resources to talk about something new. 
A key finding from these studies is that participants typically begin with verbose descriptions appealing to multiple features of the target (``a dog looking left with its tail pointed up''), but gradually form conventions and converge to a set of short labels (``the dog''). 

Classic studies have examined how newly-formed conventions transfer when the same target appears with a new \emph{social partner}~\citep{wilkes1992coordinating,metzing2003conceptual,brown2015people}, or in a new \emph{referential contexts}~\citep{brennan1996conceptual,ibarra2016flexibility}.
However, a critical axis of generalization has been comparatively underexamined: the problem of generalization to new referential \emph{targets} \citep{shepard1987toward,nolle2018emergence,raviv2022variability}.
If interlocutors are not simply memorizing a one-to-one mapping but achieving some form of broader \emph{conceptual alignment} \citep{clark1996using,stolk2016conceptual}, then we should observe a gradient of generalization to nearby locations in the same conceptual space.
Probabilistic accounts of convention formation \citep[e.g.][]{hawkins2022partners} naturally capture this gradient via hierarchical inference.
If people update their beliefs about whether a given referring expressions will be understood for the \emph{same target} in the future, they should also (more weakly) update their beliefs about other targets in the same distribution.

\begin{figure*}[t!]
\centering
\includegraphics[width=0.99\linewidth]{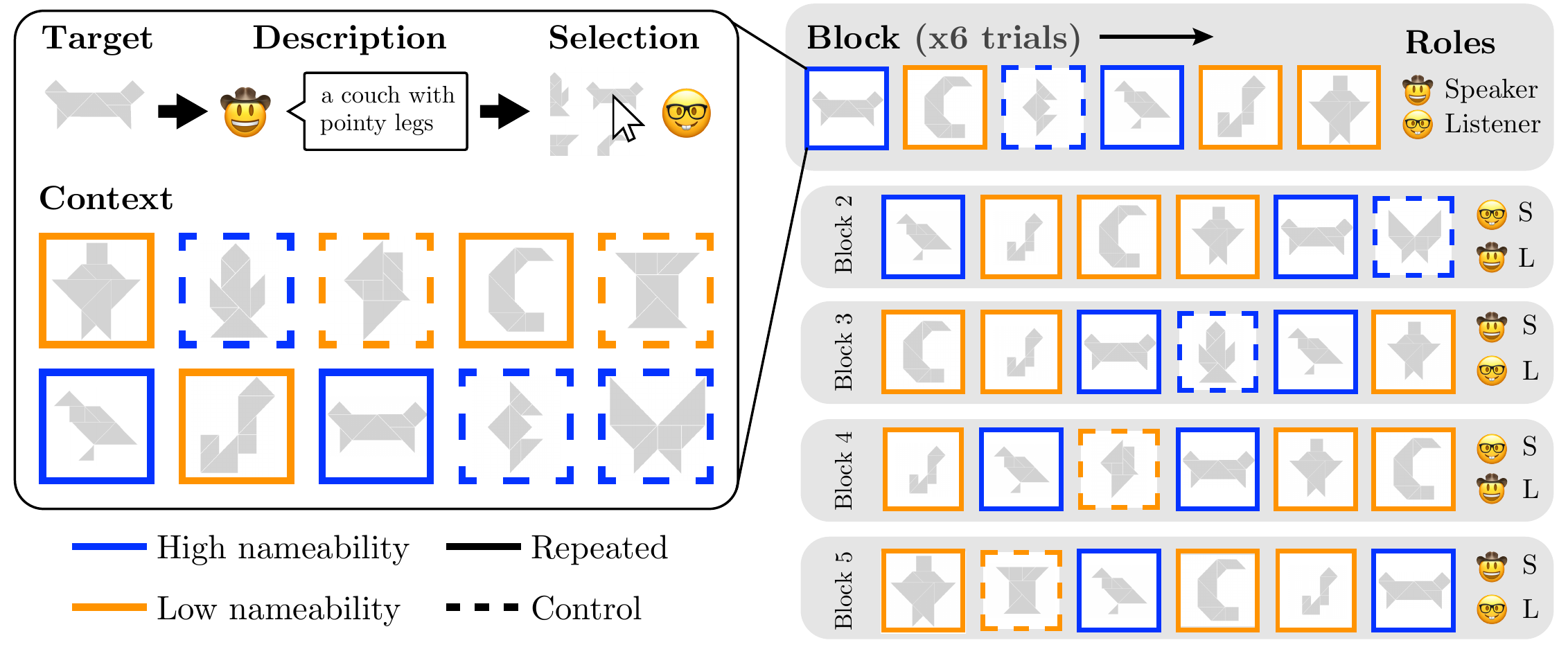}
\caption{\textbf{Experiment 1 design}. On the left, we show an example of a single reference game trial, using a mixed context of ten tangrams (borders added for this graphic). The context remains the same for all blocks and trials. On the right, we show a full target sequence consisting of five blocks. In each block, we randomly inserted a single control tangram target (dashed border) among the repeated targets (solid border).}
\label{fig:exp1}
\end{figure*}

We approach this problem by considering the property of \emph{nameability} --- roughly, the \emph{a priori} likelihood that two individuals will prefer the same label before interacting~\citep{hupet1991effects, zettersten2020finding}.
For a higher-nameability referent, such as the left image in \autoref{fig:introtangrams}, most speakers will be expected to extend the same familiar label \emph{dog}.
Meanwhile, for a lower-nameability referent, such as the image on the right, different speakers are expected to attempt to extend very different labels (e.g., ``lizard'', ``robot hand'', or even ``lighter''). 
A recent study in the color domain has suggested that speakers make fairly accurate predictions about whether a given label will be shared by others \citep{murthy2022shades}, although this may not be the case for all domains \citep{lupyan2023hidden,marti2021latent,wang2021idiosyncratic}.

In this paper, we investigate the effect of nameability on referential generalization in two studies, drawing on the recently introduced \textsc{KiloGram} collection of well-normed tangram stimuli. 
Our studies employ the well-studied repeated reference game scenario~\citep{clark1986referring}, where two participants repeatedly communicate about a set of abstract images. 
On each trial, one participant, who is assigned the speaker role, describes a prespecified target image from a set of image for the other participant, who is the listener, to select. 
Success is measured as the listener correctly selecting the target image. 
The first study focuses on the effect of nameability when reference targets repeat and conventions are formed, while the second studies the transfer of such conventions to new target images. 
The studies advance our prior understanding of generalization in reference along multiple aspects:
(a) we extend the study of repeated reference games to a much larger set of abstract tangram shapes with a higher range of nameability; 
(b) we evaluate the impact of nameability on the formation of conventions;
and (c) we measure the impact of transfer of conventions to new reference targets within a new context. 
\section{Experiment 1: Manipulating nameability}

\paragraph{Participants}
We recruited 60 pairs of participants from Prolific, based on preregistered inclusion criteria (English as first language and location based in US or UK). We excluded 8 pairs of participants, because their games contained more than 20\% empty responses. Participants provided informed consent in accordance with the institutional IRB. Each game lasted an average of 23 minutes and participants were given a base payment of \$4.25 (approximately \$11 per hour) with a performance bonus up to \$0.90.

\paragraph{Stimuli}
We designed a reference game using the black-and-white tangram shapes from the \kilogram dataset~\citep{Ji2022AbstractVR}. 
Each shape in \kilogram was previously normed for nameability using a metric called Shape Naming Divergence (SND), computed over naming annotations included in the dataset. 
This metric is defined as the mean proportion of words in each description that do not appear in any other description for that tangram. 
For example, if all annotators of a tangram used the one-word description ``bird'', the SND would be 0 because there are no unique words; if all annotators used distinct one-word descriptions, the SND would be 1. 
We sorted all 1016 tangrams in the dataset by SND and selected the top 100 tangrams to use for the \emph{low-nameability} condition and the bottom 100 for the \emph{high-nameability} condition to ensure maximal differentiation.

For each pair of participants, we sampled a context of 10 tangrams, 5 tangrams from the high-nameability set and 5 tangrams from the low-nameability set.
We ensured that tangrams in the context are sufficiently distinct to avoid challenging informativity pressures (e.g. two high-nameability tangrams that both have the consensus label \emph{bird}).
For each tangram, we extracted the head word from each of the 10 \textsc{KiloGram} descriptions using SpaCy v3 (\cite{spacy}) dependency parser.
We then set a threshold of $10\%$ overlap in any pairwise list of head words.
We reject and re-sample sets with pairs above the threshold. 

\paragraph{Design and procedure}

\autoref{fig:exp1} illustrates the design. Each game contained 5 blocks, and each block had 6 trials. All trials in a game were based on the same context. 
In each context, 5 tangrams were assigned to the \emph{repeated} condition, each appearing exactly once in each block. Among the 5 repeated tangrams, we ensured that 2 were drawn from the low-nameability condition and 3 from the high-nameability condition, or vice versa. The other 5 tangrams were assigned to the \emph{control} condition and only appear in one of the blocks as the target. 
Thus, all experimental manipulations were within-dyad. 
The order of targets are randomized in each block, and participants alternated roles between blocks. 

Participants were randomly assigned to pairs after providing consent and passing a tutorial and a quiz. 
The participants were randomly assigned to the speaker or the listener roles.
When the game started, both players saw 10 tangrams and a chat box. 
The order of the tangrams was different from the speaker's and the listener's views, so that the speaker could not rely on the position when describing the target. The speaker was asked to send only one message to describe the highlighted target tangram from the context in 45 seconds. The listener needed to select a tangram based on the description. An additional 15 seconds were given to the listener to make the selection if they had not already. Both participants received feedback after each trial indicating if the listener had responded correctly. 
At the end of the experiment, the participants were given a demographic survey about their age, gender, language background, game experience, feedback for the study, etc.
The experiment was built with Empirica~\citep{Almaatouq2020EmpiricaAV}, a platform for building and conducting synchronous and interactive online experiments with human participants. 
We hosted the games on Meteor Cloud and stored game data in MongoDB. 

\subsection{Results}

We tested three key hypotheses about performance over time. 
First, we expected that high-nameability targets would be ``easier'' to communicate about than low-nameability targets across the board.
Second, we expected that performance would improve overall for repeated targets to a greater degree than the non-repeated controls interspersed throughout the trial sequence.
Third, and of greatest theoretical interest, we ask whether there is a three-way interaction: the performance improvements that accrue over successive interactions for repeated targets may more readily transfer to improved performance on low-nameability controls than high-nameability controls.

We evaluated these predictions using a series of mixed-effects regression models for three complementary metrics of communication performance: referential accuracy (whether or not the listener successfully selected the target), verbosity (the number of words in the speaker's description), and speed (the time taken for the listener to make a selection after receiving the message).
For each metric, we construct a regression model including fixed effects of condition (repeated vs. control) and nameability class (high vs. low), as well as an effect of block number (integers 1 to 5, centered) and all their interactions.
For accuracy (coded as a binary variable: correct vs. incorrect), we use a logistic linking function. 
Because all manipulations were within-dyad, we included the maximal random effect structure at the dyad-level.

\begin{figure}[t]
\centering
\includegraphics[width=7.5cm]{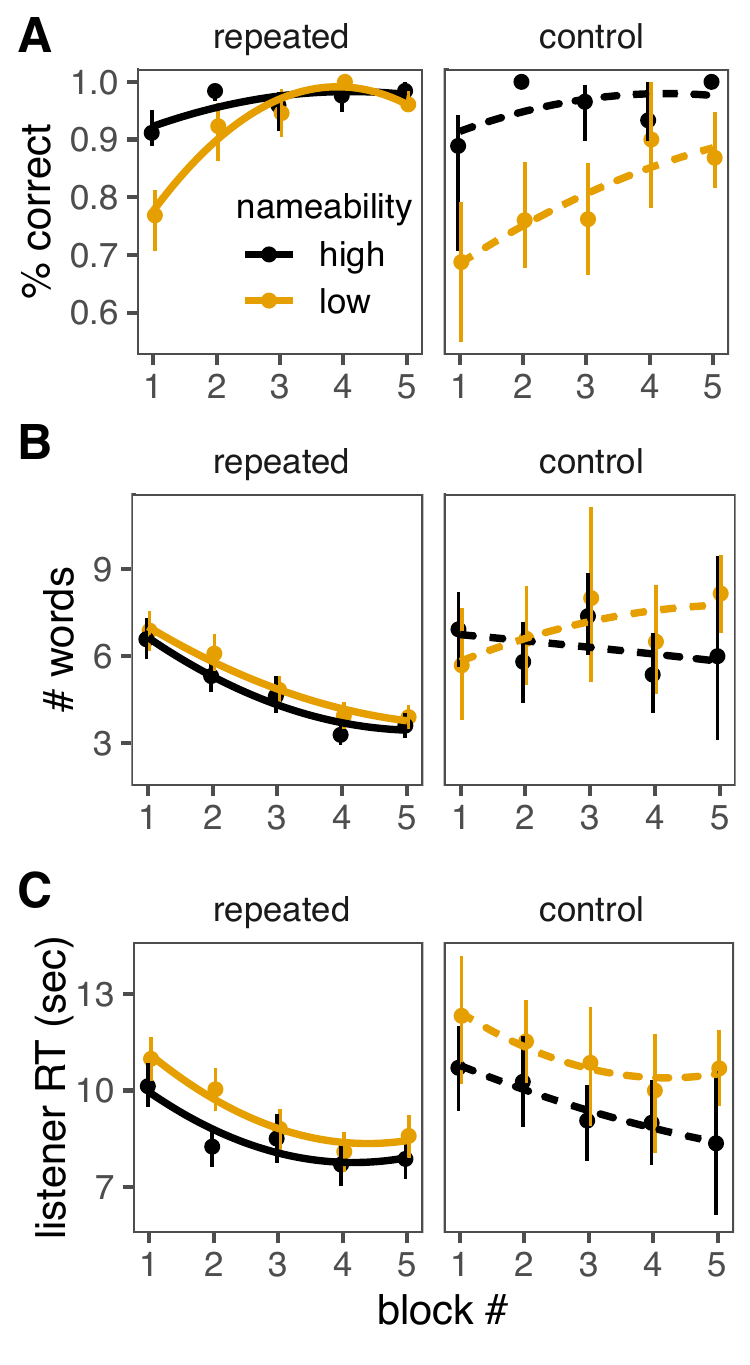}
\caption{\textbf{Experiment 1 results}. (A) accuracy, (B) description length, and (C) time elapsed between speaker message and listener response (in seconds) for low- and high-nameability tangrams in the repeated (left) or control (right) condition. Error bars are bootstrapped 95\% CIs.}\label{exp1_results}
\end{figure}

\paragraph{Accuracy}
Our raw accuracy metric is shown in \autoref{exp1_results}A.
First, we observed a significant main effect of nameability: all else equal, listeners were more likely to make errors for low-nameability targets than high-nameability ones, $b=0.87, z=-3.4, p < 0.001$.
While success rates improved overall over the course of the task, $b = 33.3, z=4.1, p < 0.001$, the most complex model supported by our data only included a significant interaction between condition and nameability, $b=-0.32, z=-2.06, p = 0.039$, likely reflecting ceiling effects for high-nameability targets, and no significant interaction was found with block number.

\paragraph{Description length}
We considered the efficiency with which speakers were able to communicate, measured as the number of words in their descriptions (\autoref{exp1_results}B).
We first found an overall main effect of block, $b=-18.6,~t(105) = -3.7,~p<0.001$, consistent with classic observations~\citep{krauss1964changes,clark1986referring}, as well as a significant main effect of nameability, $b=0.38,~t(47)= 2.8,~p=0.008$, where high-nameability targets received shorter descriptions.
We also found a significant interaction between condition and block: holding nameability constant, the decrease in utterance length for repeated tangrams was significantly greater than for control tangrams, $b = 25.2,~t(689)=6.5,~p<0.001$.
Finally, we found a significant three-way interaction, $b = 8.4,~t(1066)=2.1,$ $p=0.039$, consistent with the hypothesis that dyads are able to converge on shorter descriptions for repeated tangrams regardless of their nameability, but these improvements only generalize to high-nameability control tangrams.
If anything, when speakers must refer to a low-nameability control tangram for the first time late in the game, they produce slightly longer expressions than they did earlier in the game.

\begin{figure}[t!]
\centering
\includegraphics[width=8.5cm]{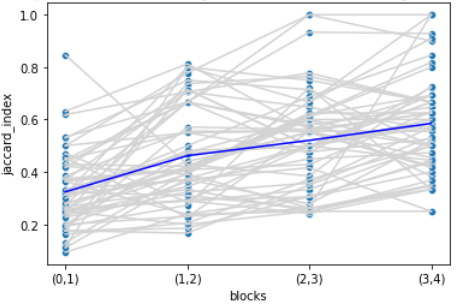}
\caption{\textbf{Experiment 1 analysis}. To evaluate the stability of referring expression content over the course of the game, we computed the Jaccard index between the set of all words used in each pair of successive blocks. We found a gradual increase in the metric, indicating increasing block-to-block similarity in word sets.} 
\label{ex1_edit_distance}
\end{figure}

\paragraph{Listener response time}
Our third performance metric is the time required for listeners to respond after receiving a description (\autoref{exp1_results}C), as listeners may be expected to pause longer when uncertain about their response.
To ensure that listener response times are not driven by the time window left by speakers, we capped response times at the maximum value of 15 seconds given to listeners on trials when the speaker ran out the clock.
We found a significant main effect of nameability, where listeners responded more quickly overall for high-nameability targets, $b=0.63,~t(139) = 4.7,$ $p<0.001$.
As with accuracy, we did not find any significant interactions with block number, but did find an interaction between condition and nameability, $ b = 0.28,~t(1372) = 2.3, p = 0.02$.
Although listeners were always a bit slower to respond for control tangrams than repeated tangrams, the gap was significantly bigger for low-nameability ones. 

\paragraph{Stability of descriptions}

So far we have observed that dyads are able to communciate more efficiently and accurately over the course of a game, as a function of nameability.
However, these coarse metrics are agnostic to the actual linguistic content of descriptions, the vocabulary used by the speaker in order to describe targets. 
To examine whether interlocutors are truly converging on shared conventions (despite swapping roles each block), we calculated a measure called the Jaccard index, $J(W_i, W_{i+1}$), between the set of words $W_i$ used on successive blocks for tangrams in the \emph{repeated} condition.
The Jaccard index is a measure of set similarity defined as the size of the intersection over the size of the union, e.g. $$J(W_1, W_2) = |W_1 \cap W_2| / |W_1  \cup W_2|$$
To ensure effects were not driven by spurious changes in function words or pluralization, we lemmatized all descriptions and excluded stop words prior to computing sets.
The results of this analysis are shown in 
We found a steady increase in similarity across successive blocks \autoref{ex1_edit_distance}, indicating that participants increasingly reused lexical choices from their partner in the previous block as  shared conventions stabilized.\footnote{We found a similar effect for a directed variant of the Jaccard index: $J_{dir}(W_1, W_2)=|W_1 \cap W_2| / |W_2|$ normalizing by the size of the later description rather than the union. This supplemental analysis suggests that the observed changes in set similarity were not driven by non-stationarity in set size over time.}
\section{Experiment 2: Generalizing to unseen targets}

\begin{figure}[t!]
\centering\includegraphics[width=0.9\linewidth,clip,trim=415 1485 295 13]{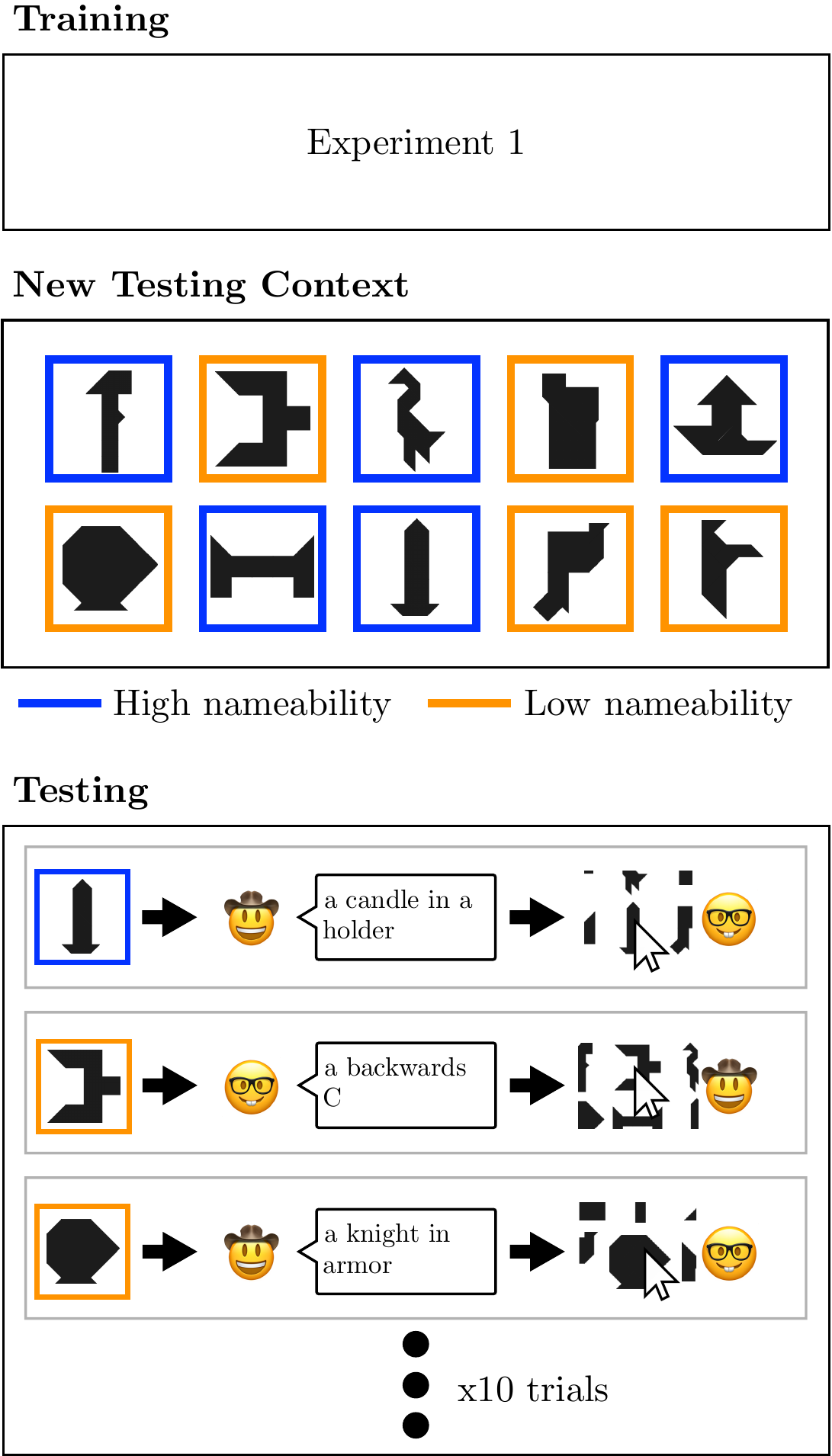}
\caption{\textbf{Experiment 2 design.} After participants completed the same procedure as Experiment 1 as a \emph{training phase}, they proceeded to a new \emph{test phase} where they played a reference game with an entirely new context of 10 tangrams.}\label{fig:exp2}
\end{figure}

Our findings from Experiment 1 not only emphasize the added difficulty of communicating about low-nameability objects overall, but also hint at the additional difficulty of \emph{generalizing} newly acquired conventions to low-nameability objects. 
This effect was particularly strong for description length, where speakers were only willing to extend reduced descriptions to high-nameability control objects.
One explanation for the differential effect of nameability on control trials is that speakers became increasingly confident that their partner would share the same meaning for unseen tangrams only when they were expected to have high consensus \emph{a priori} and existing conventions were likely to be effective \citep{murthy2022shades}.
However, it is also possible that these control tangrams were simply more \emph{familiar} as they had appeared in context alongside the repeated tangrams prior to appearing as the target.
In Experiment 2, we introduce a stronger test of this hypothesis by measuring how speakers generalize to new contexts where all targets are entirely novel.

\subsection{Methods}
\paragraph{Participants}
We recruited 60 pairs of participants, 8 of whom were excluded based on the same criteria used in Experiment 1. 
Games lasted an average of 27 minutes and participants were paid \$5.50  (approximately \$11 per hour) with a performance bonus up to \$1.20.

\paragraph{Stimuli, design and procedure}
We used the same sets of high- and low-nameability stimuli from Experiment 1, and the procedure was a direct replication and extension. 
The first phase of the experiment (the \emph{training} phase) was an exact replication of the within-dyad $2 \times 2$ design used in Experiment 1.
The second phase (the \emph{test} phase) was new.
A 6th block was appended, containing 10 additional trials (\autoref{fig:exp2}).
Critically, these test trials used a completely non-overlapping context with 10 new targets presented in a randomized sequence. 
Each tangram in the new context was given a single trial. 
Speaker and listener roles were swapped between every trial.

\subsection{Results}

We evaluated the performance of a new context by examining two metrics: accuracy and verbosity (we omit the Experiment 2 response time analysis due to space constraints).

\begin{figure}[t]
\centering
\includegraphics[width=0.99\linewidth]{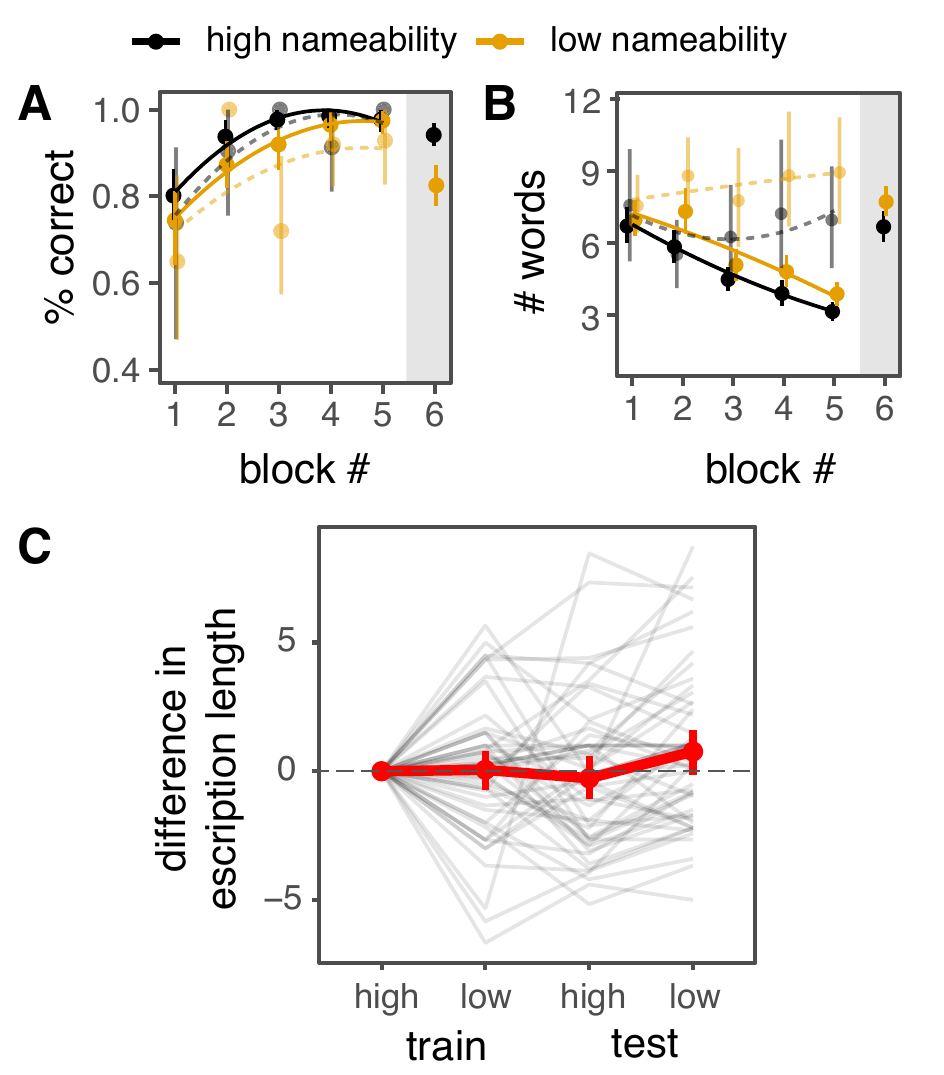}
\caption{\textbf{Experiment 2 results}. (A) Accuracy and (B) description length for the train phase (blocks 1-5) and test phase (block 6). Transparent dashed lines are control conditions. (C) Within-game differences in description length relative to the first block of train for the first train and test blocks, for high- and low-nameability targets. Error bars are bootstrapped 95\% CIs.} 
\label{exp2_results}
\end{figure}

\paragraph{Accuracy}

In addition to replicating the nameability effects examined in Experiment 1 (see \autoref{exp2_results}A), the primary analysis of interest is a direct comparison against the initial block of the train phase (block 1) and the initial block of the test phase (block 6).
In both cases, it is the speaker's first time referring to all targets in context.
Thus, any differences in the test phase can be attributed to some generalizable learning taking place over the training block.
We ran a mixed-effects logistic regression model predicting accuracy only for these two blocks, including fixed effects of phase (train vs. test) and nameability (low vs. high), as well as random intercepts and slopes at the dyad-level.
We found a significant interaction, $b=-0.31, z=2.25, p=0.024$, indicating that there was greater improvement from the beginning of train to test for high-nameability tangrams than low-nameability ones.

\paragraph{Description length}
Average description lengths are shown for the training phase and test phase in \autoref{exp2_results}B. 
We again replicated the nameability effects found in Experiment 1, and focus here on the direct comparison between the first block of training and the first block fo test. 
We ran a mixed-effects linear regression model predicting description length, including fixed effects of phase (train vs. test) and nameability (low vs. high) with random intercepts and slopes at the dyad-level.
In addition to a significant main effect of nameability at the beginning of both train and test, $b=0.32, t(112)=-2.181, p=  0.03$, we found a marginal interaction, $b=0.25, t(629)=1.8, p = 0.069$. 
In a Bayesian mixed-effects model with full random effects, we obtained a 95\% credible interval of $[-0.05, 0.54]$. 
In other words, there was  a small but meaningful gap in description length between high and low-nameability tangrams in the test phase, while no such gap was observed in the first block (see \autoref{exp2_results}C for a finer-grained visualization controlling for individual differences in overall verbosity). 
Put together, these effects suggest that participants were better able to anticipate in the test phase which tangrams would be harder and adjust their description length accordingly.

\section{Discussion}

The study conducted in this paper investigated the effect of nameability on the process of reference generalization using tangrams from the \kilogram dataset, a large-scale high-diversity resource for abstract stimuli. 
We conducted  two experiments that track the development of conventions during a repeated reference game, and assessed the impact of these conventions on a reference game with a new context. 
We observed that conventions, expressed through more efficient communication and stable vocabulary, are formed through repeating games, re-affirming observations from past work~\citep{clark1986referring,hawkins2020characterizing} using the larger-scale \kilogram data \citep{Ji2022AbstractVR}. 
Our key contributions concern the transfer of conventions to novel stimuli (i.e. control images), and the influence of nameability on this process. 
Our analyses showed that conventions formed for repeating stimuli transfer to both control stimuli that do not repeat, and to completely new context with previously unseen tangrams. 
This form of generalization has been generally understudied in prior work. 
Moreover, this process is modulated by the nameability of the target objects. 
Objects with higher nameability lent themselves to transfer of acquired conventions significantly more than objects with lower nameability.

Our studies raise several important directions for future work. 
Most importantly, we leave open the question of what, exactly, is transferred. 
One possibility is that pairs of participants aligned their semantic spaces in a way that generalized beyond the specific visual stimuli observed in the repeated condition.
This conceptual alignment then allowed them to produce and comprehend references in new contexts more efficiently and more effectively. 
The differences we observed across nameability classes suggest that this alignment is not perfect, and generally noisier for objects that are harder to describe. 
Indeed, the conceptual alignment hypothesis predicts that generalization would fall off as a function of perceptual or conceptual distance.
An alternative hypothesis is that participants developed a shared model of which stimuli are easy or hard to describe (i.e. have high or low nameability). 
The speakers could have then utilized this information and the mutual knowledge that it is shared with their partners to be more efficient referring to objects in the new context. 
In a nutshell, the two hypotheses contrast the direct tuning of semantic knowledge against the acquisition of a higher-level task model. 
Identifying which, if either, of these hypotheses is correct requires further studies, likely with more constrained structure. For example, rather than using behavior during the first block as a baseline, which also impacts the actual conventions formed, future studies could prefix the repeating game with an asocial naming block. 

Another important direction for future work concerns how the statistics of the object distribution in each of the contexts (either for the train or test phase), may influence the transfer of conventions. 
For example, could contexts with a larger ratio of high nameability objects facilitate faster convention formation? 
Could contexts with more low nameability objects provide a more challenging environment and result in more enduring conventions? 
Our studies have also not explored the impact of the semantics of the objects on participants behavior. 
For example, if we were to create contexts that include more semantically related objects, would that lead to faster or slower convention formation? 
Would it result in more robust transfer to new contexts? 
Many of these questions could be readily addressed with carefully designed contexts making use of the diversity and scale of \kilogram.

It is highly unlikely for the exact same concepts and situations to repeat in realistic situations.
Thus, generalizing partner-specific conventions to new concepts is an crucial part  streamlining interaction with repeating partners. 
Understanding and modeling such generalization behavior is paramount to gain insight into the development of language as a system for coordination and efficient communication between individuals. 

\section{Acknowledgments}

This research was supported by ARO W911NF21-1-0106 and NSF grant 1750499 to YA. 
RDH was supported by a C.V. Starr Fellowship.

\printbibliography

\end{document}